\documentclass[letterpaper, 10 pt, conference,onecolumn]{ieeeconf}  

\IEEEoverridecommandlockouts                              

\usepackage{cite}
\usepackage{amsfonts}  
\usepackage{pdfpages}
\usepackage{amsmath}
\usepackage{booktabs}
\usepackage{multirow}
\usepackage{caption} 
\usepackage{pdfpages}
\usepackage{graphicx}
\usepackage{subcaption}
\usepackage{capt-of}
\usepackage[T1]{fontenc}
\usepackage{aecompl}
\usepackage{hyperref}
\title{\LARGE \bf
Structure-aware Hypergraph Transformer for Diagnosis Prediction in Electronic Health Records* 
}

\author{Haiyan Wang and Ye Yuan, \textit{Member}, IEEE
\thanks{*This research is supported by the National Natural Science Foundation of 
	China under grants 62372385 (Corresponding Author: Y. Yuan).}
\thanks{H. Y. Wang and Y. Yuan are with the College of Computer and Information 
	Science, Southwest University, Chongqing 400715, China ({e-mail:  why040680@swu.edu.cn}, yuanyekl@swu.edu.cn)}}

\begin{document}

\maketitle
\thispagestyle{empty}
\pagestyle{empty}

\begin{abstract}
Electronic Health Records (EHR) systematically organize patient health data through standardized medical codes, serving as a comprehensive and invaluable resource for diagnosis prediction. Graph neural networks (GNNs) have demonstrated effectiveness in modeling interactions between medical codes within EHR. However, existing GNN-based methods face two key limitations when processing EHR: a) their reliance on pairwise relations fails to capture the inherent higher-order dependencies in clinical data, and b) the localized message-passing mechanisms extract global code interactions inadequately. To address these issues, this paper proposes a novel Structure-aware HyperGraph Transformer (SHGT) framework following three-fold ideas: a) modeling EHR data as a hypergraph and employing a hypergraph structural encoder to capture higher-order interactions among medical codes, b) integrating the Transformer architecture to effectively capture global dependencies across the entire hypergraph, and c) designing a tailored loss function incorporating hypergraph reconstruction to preserve the hypergraph's original structure. Extensive experiments on two real-world EHR datasets demonstrate that the proposed SHGT outperforms existing state-of-the-art models on diagnosis prediction.
\end{abstract}

\section{INTRODUCTION}

Driven by the rapid evolution of information technology and the widespread implementation of Hospital Information Systems (HIS), the accessibility of Electronic Health Records (EHR) has substantially improved \cite{sarwar2023secondary,bohr2020rise,ref29,ref30,ref31}. By integrating comprehensive patient healthcare data---such as diagnosis codes, medication codes, and procedure codes---EHR have drawn significant research interest \cite{ref33,ref34,ref35,ref36}, prompting extensive studies on these digitized medical records to advance intelligent healthcare \cite{pieroni2021predictive,ref37,ref38}, particularly in clinical decision support tasks such as diagnosis prediction \cite{ref39,ref40,cui2022medskim,ref32}, medication recommendation \cite{li2022knowledge,ref41,ref42,ref43}, and risk assessment \cite{ref44,ref45,ye2021medpath,ref46,tariq2023fusion,ref47}.

In recent years, deep learning methodologies have been widely adopted for predicting diagnoses and optimizing clinical decision-making \cite{ref48,xiao2018deep,yin2019deep,yang2023pyhealth}. For instance, traditional sequence-based methods \cite{ref49,choi2016retain,ref50} represent a patient's historical visits as strictly ordered sequences, capturing temporal dependencies and variations across visits. However, within a single visit, the medical codes do not follow a clear temporal order, posing challenges for these models in accurately capturing complex interactions among medical codes, thereby limiting their representational capacity. 

Graph Neural Networks (GNNs) \cite{kipf2017semi,ref1,ref2,ref3} have recently gained significant attention for their remarkable advancements in healthcare, owing to their superior representational capability \cite{tang2023readmission,ref4,ref5}. Existing GNN-based approaches \cite{ref8,wei2022dynamic,ref9} decompose each visit into pairwise code relationships, constructing code-code graphs based on their co-occurrence within individual visits. Despite capturing interactions between medical codes, these methods face the following limitations: a) \textit{\textbf{Limited Capture of Higher-Order Interactions:}} These methods typically construct pairwise code graphs based on local co-occurrence patterns, making it difficult to capture complex higher-order interactions among medical codes within a single visit. This limitation undermines the holistic structural representation of clinical visits. b) \textit{\textbf{Insufficient Global Contextual Awareness:}} These approaches rely on localized message-passing mechanisms to learn code embeddings. However, such strategies overlook the global dependencies among medical codes across the entire graph, leading to less informative visit-level embeddings and suboptimal predictions.

To address the aforementioned challenges, we propose a \textbf{S}tructure-aware \textbf{H}yper\textbf{G}raph \textbf{T}ransformer for diagnosis  prediction, named SHGT. Unlike traditional GNNs, which typically construct code-code graphs based on the co-occurrence of diagnosis codes in visits, SHGT utilizes a hypergraph to model EHR, capturing higher-order interactions among diverse medical codes, including diagnoses, medications, and procedures. Additionally, we introduce a Hypergraph Structural Encoder to capture the local structural information of medical codes and visits, which is then fed into the Transformer to model global interactions. To preserve the original structure of the hypergraph, we design a novel loss function that incorporates hypergraph reconstruction loss, ensuring semantic alignment between medical codes and their corresponding visits. To summarize, the main contributions of this work are as follows:
\begin{itemize}
	\item We present a novel model, SHGT, which represents EHR data as a hypergraph to capture higher-order interactions among various types of medical codes. In addition, we introduce a hypergraph structural encoder to extract local structural information, which is subsequently injected into the Transformer architecture, facilitating the learning of global code dependencies within the hypergraph.
	\item We propose a novel loss function combining hypergraph reconstruction loss with predictive loss, ensuring hypergraph structural integrity during optimization and mitigating the loss of structural information.
	\item Comprehensive empirical evaluations are conducted on two clinical datasets, with comparative results against seven state-of-the-art methods demonstrating the superior effectiveness of SHGT.
\end{itemize}

\section{RELATED WORK}

\subsection{Traditional Sequence-Based Diagnosis Prediction}

Sequence-based models effectively capture temporal patterns in EHR by organizing patient visits into ordered sequences, where each visit reflects a patient's medical status at a given time \cite{ref21,ref22}. For example, Doctor AI \cite{choi2016doctor} employs a recurrent neural network (RNN)-based approach to capture temporal dependencies in patient records, enabling it to model the sequence of medical events over time and make predictions based on historical health data. RETAIN \cite{choi2016retain} combines an RNN with a two-level attention mechanism, thereby enhancing prediction accuracy and improving model interpretability. StageNet \cite{gao2020stagenet} applies stage-aware long short-term memory (LSTM) to capture disease progression and employs a convolutional module to incorporate disease development patterns into risk prediction. HiTANet \cite{luo2020hitanet} leverages Transformer to embed temporal information into visit-level representations and assigns global weights across time steps to generate patient representations. CEHR-BERT \cite{619472d55244ab9dcbd2de8d} modifies the BERT architecture by incorporating timestamp embeddings to handle irregular time intervals, while leveraging contextual information to generate more comprehensive patient representations. Despite their advantage in capturing temporal associations between visits, they fail to account for intra-visit relationships among medical codes, which are crucial for diagnosis prediction tasks\cite{ref10,ref11,ref12,ref13}. 

\subsection{Graph-Based Diagnosis Prediction}

 Unlike sequence-based approaches, graph-based methodologies conceptualize the EHR as a graph, inherently preserving its rich structural information \cite{ref14,ref15,10831750,ref16, wang2024faircare,ref17}. For instance, GRAM \cite{choi2017gram} constructs a knowledge graph of medical codes based on a medical ontology and employs graph attention networks to learn representations of the codes. GCT \cite{choi2020graphical} creates a comprehensive medical graph that integrates diagnoses, treatments, and laboratory results. By combining graph convolutional networks and transformers, it explores the hidden structure of EHR. T-ContextGGAN \cite{9794568} designs a graph attention network with time-aware meta-paths along with self-attention mechanisms to simultaneously extract temporal semantic information and inherent relationships. BioDynGraph \cite{li2024biodyn} leverages the co-occurrence relationships of diagnostic codes across patient visits to formulate a global disease co-occurrence graph, which is then learned through an event-based clue search mechanism. GraphCare \cite{jiang2024graphcare} leverages large language models to extract knowledge from external biomedical knowledge graphs, generating patient-specific knowledge graphs. These knowledge graphs are then used to train a dual-attention-enhanced BAT graph neural network for medical prediction.  Despite their performance gains, these models remain limited to pairwise code-level interactions, failing to capture the complex higher-order dependencies among heterogeneous medical codes\cite{ref18,ref19,ref20}.

\section{PRELIMINARY}
\subsection{Medical Codes and Visits} In the context of EHR, the complete set of medical codes is denoted by $C= \{c_1, c_2, \ldots, c_m\}$, where $m$ denotes the total number of unique medical codes. Each medical code $c$ belongs to  one of three categories: diagnosis codes, medication codes, and procedure codes, reflecting different aspects of patient care. Specifically, diagnosis codes are represented by $D = \{d_1, d_2, \ldots, d_{|D|}\}$, where $|D|$ denotes the total number of diagnosis codes. Since each patient visit $v$ corresponds to an inpatient admission characterized by a diverse set of medical codes, it can be formally structured as a collection of these codes.
\subsection{EHR Dataset}

The EHR dataset comprises comprehensive hospitalization records for all patients, formally defined as 
$P = \{p_\mathit{u} \mid u \in U\}$, 
where $U$ represents the set of all patients. For each patient $p_u$, the visit history is expressed as 
$p_{\mathit{u}} = \{v_{\mathit{u}}^1, v_{\mathit{u}}^2, \ldots, v_{\mathit{u}}^T\}
$, 
with $T$  as the total number of visits for the given patient. Across the entire dataset, the total number of visits is denoted as $n$, reflecting the overall scale of documented admissions.

\subsection{Diagnosis Prediction}
For a patient $p$ with $T$ prior visits, the objective of diagnosis prediction is to predict the set of diagnoses expected in the next visit. It formally involves estimating the likelihood of each diagnosis in the $(T+1)$-th visit, formulated as 
$y^{T+1} \in \{0, 1\}^{|D|}$.

\section{METHODOLOGY}
\begin{figure*}[!t]
	\centering
	\includegraphics[width=\textwidth]{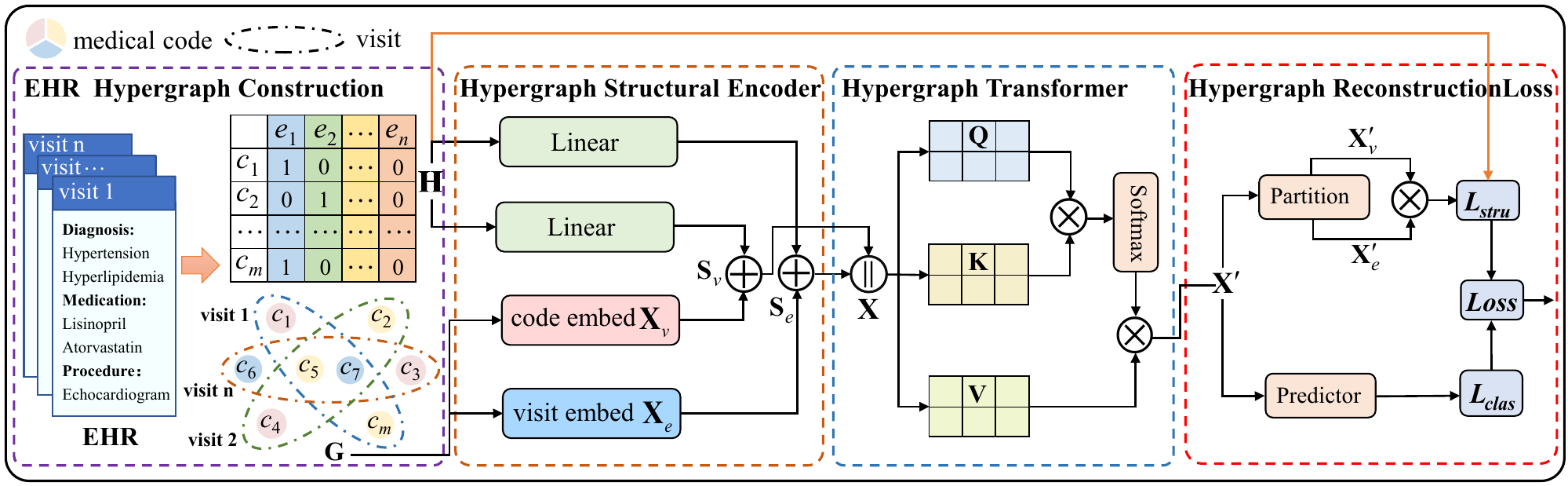}
	\caption{The overall architecture of our SHGT}
	\label{model}  %
\end{figure*}
This section provides a detailed overview of the proposed SHGT, with Fig. \ref{model} showing its overall architecture. We begin by detailing the construction of a hypergraph from EHR data. Subsequently, we elaborate on how the hypergraph structural encoder captures local structural information, while the Transformer models global interactions among medical codes across the entire hypergraph. Finally, we incorporate hypergraph reconstruction loss into diagnosis prediction and present the optimization strategy.

\subsection{EHR Hypergraph Construction}

To effectively capture complex higher-order interactions among diverse types of medical codes, we employ a hypergraph-based representation. In contrast to methods relying on vanilla graphs, hypergraphs enhance connectivity by simultaneously linking multiple nodes. Specifically, by encoding medical codes as nodes and representing each visit as a hyperedge, we construct a hypergraph that not only captures intricate higher-order interactions among medical codes but also explicitly retains the associations between medical codes and visits. Formally, the constructed hypergraph is denoted as $G = (V, E, \mathbf{H})$, where $V$ represents the set of nodes corresponding to the complete set of medical codes, consisting of $m$ elements, and $E$ denotes the set of hyperedges, with cardinality $n$, corresponding to the number of visits. The hypergraph structure is characterized by an incidence matrix $\mathbf{H} \in \mathbb{R}^{m \times n}$, where each entry $\mathbf{H}_{ij}$ is assigned a value of $1$ if medical code $c_i$ is present in visit $e_j$, and $0$ otherwise, as illustrated in Fig. \ref{model}.

\subsection{Hypergraph Structural Encoder}
Local graph structures have demonstrated their effectiveness in graph representation learning \cite{xia2022selfsupervised,liu2024hypergraph}. Consequently, we leverage the incidence matrix $\mathbf{H}$ to capture localized code-code interactions and code-visit associations, as it inherently encodes the topological structure of the hypergraph. The process begins with the initialization of the medical code embedding matrix $\mathbf{X}_v \in \mathbb{R}^{m \times d}$. To ensure semantic alignment between visits and medical codes, we apply a mean pooling operation to aggregate the embeddings of medical codes associated with each visit, thereby generating the visit embedding matrix
 $\mathbf{X}_e \in \mathbb{R}^{n \times d}$. The localized structural embeddings for medical codes and visits are then computed as follows:
\begin{equation}
	\mathbf{S}_v = \mathbf{H} \mathbf{W}_v \in \mathbb{R}^{m \times d}
\end{equation}
\begin{equation}
	\mathbf{S}_e = \mathbf{H}^{\mathrm{T}} \mathbf{W}_e \in \mathbb{R}^{n \times d}
\end{equation}
where $\mathbf{W}_v$ and $\mathbf{W}_e$ denote learnable transformation matrices. We then obtain the structure-aware code embedding $\mathbf{Z}_v$ and visit embeddings $\mathbf{Z}_e$ by fusing structural information with the original embeddings through element-wise addition, formulated as:
\begin{equation}
	\mathbf{Z}_v = \mathbf{S}_v + \mathbf{X}_v
\end{equation}
\begin{equation}
	\mathbf{Z}_e = \mathbf{S}_e + \mathbf{X}_e
\end{equation}
\subsection{Hypergraph Transformer}
In practical medical scenarios, interactions among medical codes are not confined to a single visit. For instance, a diabetes diagnosis recorded in an initial visit may later lead to kidney disease, which is only detected in the third visit. Additionally, a portion of the medical codes recorded within each visit may be irrelevant to the diagnosis prediction task. Consequently, filtering out these irrelevant nodes is crucial for generating meaningful visit embeddings. To address these challenges, we leverage the self-attention mechanism from the Transformer architecture to automatically capture the relevance between any pair of elements, enabling dynamic modeling of complex cross-visit interactions. 

We integrate the code embeddings and visit embeddings as inputs, formally expressed as:
\begin{equation}
	\mathbf{X} = \begin{bmatrix}
		\mathbf{Z}_v \\
		\mathbf{Z}_e
	\end{bmatrix}
\end{equation}
where $\mathbf{X}$ represents the input embedding matrix. We then compute the attention matrix to capture global relationships between medical codes and visits, which is defined as:
\begin{equation}
	\textbf{A}^{(\ell)} = \mathrm{softmax}\left( \frac{\left( \mathbf{X}^{(\ell)} \mathbf{W}_Q^{(\ell)} \right) \left( \mathbf{X}^{(\ell)} \mathbf{W}_K^{(\ell)} \right)^{\mathrm{T}}}{\sqrt{d}} \right)
\end{equation}
Here, $\textbf{A}^{(\ell)} \in \mathbb{R}^{(m+n) \times (m+n)}$ encodes the interaction strengths among all elements, $d$ denotes the dimension of the embeddings. The input embeddings are projected into query and key spaces using the learnable weight matrices $\mathbf{W}_Q^{(\ell)}$ and $\mathbf{W}_K^{(\ell)}$, respectively. The softmax function is applied to normalize the attention scores. The next layer embeddings are computed as:
\begin{equation}
	\mathbf{X}^{(\ell+1)} = \mathbf{A}^{(\ell)} \mathbf{X}^{(\ell)} \mathbf{W}_V^{(\ell)} 
\end{equation}
here, ${\textbf{X}}^{(\ell+1)}$ represents the embedding matrix of the $(\ell+1)$-th layer, $\mathbf{W}_V^{(\ell)}$ is the learnable value projection matrix. 

In this manner, the model automatically captures key interactions from any code or visit to others in the hypergraph, effectively facilitating a comprehensive understanding of the hypergraph’s global structure.
\begin{table*}[t]
	\centering
	\caption{Summary of Experimental Datasets}
	\renewcommand{\arraystretch}{1} %
	\begin{tabular}{lccccc}
		\toprule
		\textbf{Dataset} & \textbf{\#Patients} & \textbf{\#Visits} & \textbf{\#Diagnoses} & \textbf{\#Medications} & \textbf{\#Procedures} \\
		\midrule
		MIMIC-III  & 5,442  & 14,124  & 1,956 & 300  & 1,399 \\
		MIMIC-IV   & 10,000 & 29,303  & 1,134 & 281  & 899   \\
		\bottomrule  
	\end{tabular}
	\label{dataset}
\end{table*}
\begin{table*}[t]
	\centering
	\caption{Comparative Results on Diagnosis Prediction (Best in Bold)}
	\begin{tabular}{lccccccccc}
		\toprule
		\multirow{2}{*}{\textbf{Models}} & \multicolumn{3}{c}{\textbf{MIMIC-III}} & \multicolumn{3}{c}{\textbf{MIMIC-IV}} \\
		\cmidrule(lr){2-5} \cmidrule(lr){5-7}
		& w-$F_1$ & R@10 & R@20 & w-$F_1$ & R@10 & R@20\\
		\midrule
		Retain & 19.23$\pm$0.21 & 24.12$\pm$0.16 & 33.72$\pm$0.25 & 26.12$\pm$0.34 & 30.13$\pm$0.38 & 41.06$\pm$0.37 \\
		Transformer & 18.81$\pm$0.36 & 23.90$\pm$0.25 & 33.24$\pm$0.31 & 25.59$\pm$0.33 & 29.90$\pm$0.36 & 40.56$\pm$0.61 \\
		StageNet & 18.32$\pm$0.11 & 22.62$\pm$0.39 & 32.29$\pm$0.44 & 22.76$\pm$0.26 & 27.80$\pm$0.43 & 38.60$\pm$0.34 \\
		CGL & 22.21$\pm$0.09 & 24.71$\pm$0.07 & 34.39$\pm$0.11 & 29.82$\pm$0.02 & 32.76$\pm$0.06 & 42.75$\pm$0.15 \\
		Chet & 22.07$\pm$0.18 & 24.92$\pm$0.15 & 34.56$\pm$0.14 & 29.83$\pm$0.13 & 34.09$\pm$0.12 & 43.95$\pm$0.09 \\
		Sherbet & 23.28$\pm$0.12 & 25.06$\pm$0.20 & 35.15$\pm$0.23 & 30.44$\pm$0.41 & 34.47$\pm$0.22 & 44.65$\pm$0.24 \\
		ADRL & \underline{23.88$\pm$0.21} & \underline{25.64$\pm$0.18} & \underline{35.30$\pm$0.13} & \underline{32.02$\pm$0.14} & \underline{34.72$\pm$0.08} & \underline{44.91$\pm$0.11} \\
		SHGT (Ours) & \textbf{25.60$\pm$0.15} & \textbf{27.16$\pm$0.11} & \textbf{37.48$\pm$0.13} & \textbf{34.29$\pm$0.19} & \textbf{36.25$\pm$0.13} & \textbf{46.77$\pm$0.16} \\
		\bottomrule
	\end{tabular}
	\label{comparison}
\end{table*}
\subsection{Hypergraph Reconstruction Loss}
While the hypergraph transformer effectively captures global dependencies among medical codes, it lacks a structural constraint to preserve the topology of the hypergraph, potentially leading to the loss of essential structural information. To mitigate this, we introduce a hypergraph reconstruction loss to enforce structural consistency by aligning learned embeddings with the hypergraph incidence matrix.
Let $\mathbf{X}'$ denote the final output of the Transformer, from which the embeddings of medical codes $\mathbf{Z}_v'$ and visits $\mathbf{Z}_e'$ are derived. To reconstruct the hypergraph topology, we approximate the incidence matrix as:
\begin{equation}
	\mathbf{H}' = \sigma\left( \mathbf{Z}_v' \mathbf{Z}_e'^{\mathrm{T}} \right) 
\end{equation}
where $\mathbf{H}'$ denotes the predicted hypergraph structure, the \( \sigma(\cdot) \)  is the sigmoid activation function. Due to the extreme sparsity of the hypergraph derived from EHR, the abundance of negative samples (\( \textbf{H}_{ij} = 0 \)) biases predictions toward zero. To mitigate this, we use negative sampling to maintain a balanced ratio of positive (\( P \)) and negative (\( N \)) samples. The binary cross-entropy loss is then optimized to align the predicted and actual hypergraph topology:
\begin{equation}
	L_{\text{stru}} = - \frac{1}{|P| + |N|} \sum_{(i,j) \in P} \log \textbf{H}'_{ij} + \sum_{(i,j) \in N} \log(1 - \textbf{H}'_{ij}) 
\end{equation}
\subsection{Prediction and Optimization}
Given that each patient typically has 2-3 visits, we obtain patient-level embeddings by applying mean pooling over their visit embeddings. These embeddings are then passed through a dense layer with a sigmoid activation function to generate predicted probabilities. The classification loss is formulated using the binary cross-entropy loss: 
\begin{equation}
	L_{\text{clas}} = - \frac{1}{|U|} \sum_{i=1}^{U} \sum_{j=1}^{m} \left( y_{ij} \cdot \log (\hat{y}_{ij}) + (1 - y_{ij}) \cdot \log (1 - \hat{y}_{ij}) \right)
\end{equation}
Here, \( y_{ij} \) denotes the ground-truth label for the \( j \)-th diagnosis of the \( i \)-th patient, and \( \hat{y}_{ij} \) is the corresponding predicted probability.
To jointly capture semantic accuracy and structural consistency, we combine the classification loss with the hypergraph reconstruction loss to form the final objective:
\begin{equation}
	L = L_{\text{clas}} + \alpha L_{\text{stru}}
\end{equation}
where \( \alpha \) serves as a coefficient to balance the contributions of the two losses.

\section{EXPERIMENTS}
\subsection{General Settings}
\textbf{Datasets.} We employ MIMIC-III \cite{johnson2016mimic3} and MIMIC-IV \cite{johnson2023mimic4} datasets to assess the performance of the proposed model. MIMIC-III contains de-identified healthcare data from the Beth Israel Deaconess Medical Center between 2001 and 2012, while MIMIC-IV covers admissions from 2008 to 2022. To enable diagnosis prediction based on historical visits, we exclude patients who have only one visit. To avoid temporal overlap, we randomly sample 10,000 patients from the processed MIMIC-IV dataset, restricting admissions to the period from 2013 to 2022. The statistical summary of the processed datasets is presented in Table \ref{dataset}.

\textbf{Evaluation Metrics.} To evaluate the diagnosis prediction task, we adopt the weighted $F_1$ score (w-$F_1$) \cite{ref23,ref24,ref25} and top-$k$\cite{ref26,ref27,ref28} recall (R@$k$) as performance metrics. Considering that each patient visit in the datasets contains an average of 14 diagnosis codes, we set $k$ to 10 and 20 in our experiments.

\textbf{Compared Models.} We evaluate our model against seven state-of-the-art baselines, including three sequence-based approaches: Retain \cite{choi2016retain}, Transformer \cite{vaswani2017attention}, and StageNet \cite{gao2020stagenet}, as well as four graph-based methods: CGL \cite{lu2021collaborative}, Chet \cite{lu2022contextaware}, Sherbet \cite{lu2023selfsupervised}, and ADRL \cite{CHENG2025103098}.

\textbf{Implementation Details.} For all models, each dataset is split into training, validation, and test sets following a 7:1:2 ratio. Each experiment is conducted five times using different random seeds, and we report the mean performance along with the standard deviation. For baseline models, we use the hyperparameters specified in the original papers. For SHGT, we conduct a grid search with a learning rate of 0.004, a dropout rate of 0.4, and an embedding dimension of 256. The hyperparameter $\alpha$ is set to 0.3 for MIMIC-III and 0.2 for MIMIC-IV to accommodate dataset-specific characteristics.
\subsection{Performance Comparison and Analysis}
The experimental results are summarized in Table \ref{comparison},  which clearly demonstrate that the proposed SHGT model outperforms existing methods on both datasets. Based on these results, we make the following key observations:

First, sequence-based models exhibit the poorest performance among all evaluated methods. This may be attributed to the inherent characteristics of clinical data, where diagnostic, medication, and procedural information often lack clear temporal patterns and show significant heterogeneity. While sequence-based models are effective for handling temporal sequences, they face challenges in capturing the complex dependencies between medical codes.

Second, graph-based models achieve improved performance over sequence-based approaches. For example, compared to RETAIN, the best-performing sequence-based model in experiments, the graph-based model CGL achieves improvements of 14.45\% and 13.81\% in w-$F_1$ on the MIMIC-III and MIMIC-IV datasets, respectively. This improvement can primarily be attributed to the advantages of graph structures in capturing complex interactions among medical codes, making them more suitable for modeling the intricate and heterogeneous nature of EHR data.

Third, the proposed SHGT model achieves superior performance, outperforming all competing methods across both datasets. Although models like CGL, Chet, Sherbet, and ADRL perform relatively well, they are limited to capturing only pairwise interactions between medical codes. In contrast, SHGT utilizes a hypergraph structure to model EHR data, employing hyperedges to simultaneously encode interactions among multiple medical codes, thereby effectively capturing higher-order interactions. Additionally, the integration of the Transformer architecture facilitates the modeling of global code dependencies. This richer structural representation allows SHGT to generate more expressive visit-level and patient-level embeddings, ultimately yielding outstanding performance.
\subsection{Ablation Study}
In this section, we conduct a series of ablation studies to assess the contributions of each component within the SHGT model. Specifically, we evaluate SHGT alongside its three variants: SHGT-w/o-S, which excludes the hypergraph structure encoding; SHGT-w/o-T, which removes the Transformer architecture; and SHGT-w/o-L, which omits the hypergraph reconstruction loss. 

\begin{figure}[h]
	\centering
	\includegraphics[width=0.75\linewidth]{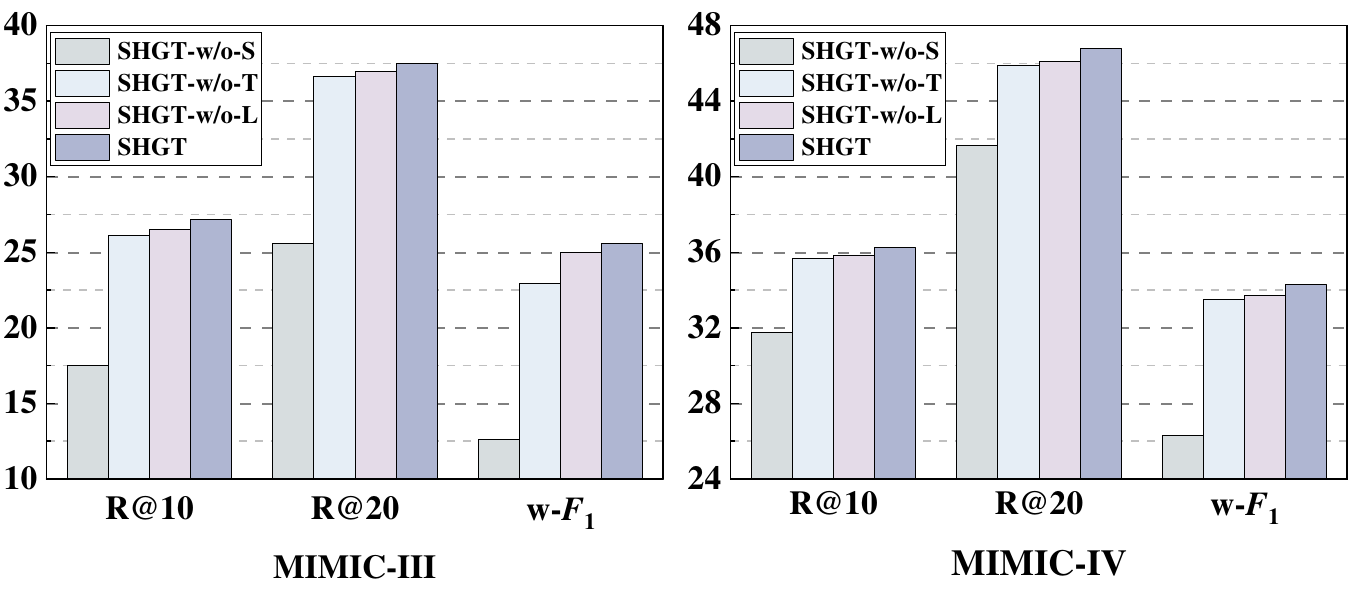}
	\caption{Ablation experiments of variant methods}
	\label{aba}  
\end{figure}

The experimental results across the two datasets are presented in Fig. \ref{aba}. As anticipated, the performance of SHGT-w/o-S declines significantly across both datasets, highlighting the critical importance of capturing higher-order interactions among medical codes for diagnosis prediction. Additionally, SHGT-w/o-T exhibits inferior performance, emphasizing the necessity of modeling global interactions among medical codes to better understand EHR data and generate informative embeddings. Finally, SHGT-w/o-L also performs worse than the full model, demonstrating the necessity of preserving structural integrity via reconstruction loss.
\begin{figure}[!t]
	\centering
	\begin{subfigure}[b]{0.43\columnwidth}
		\centering
		\includegraphics[width= \linewidth]{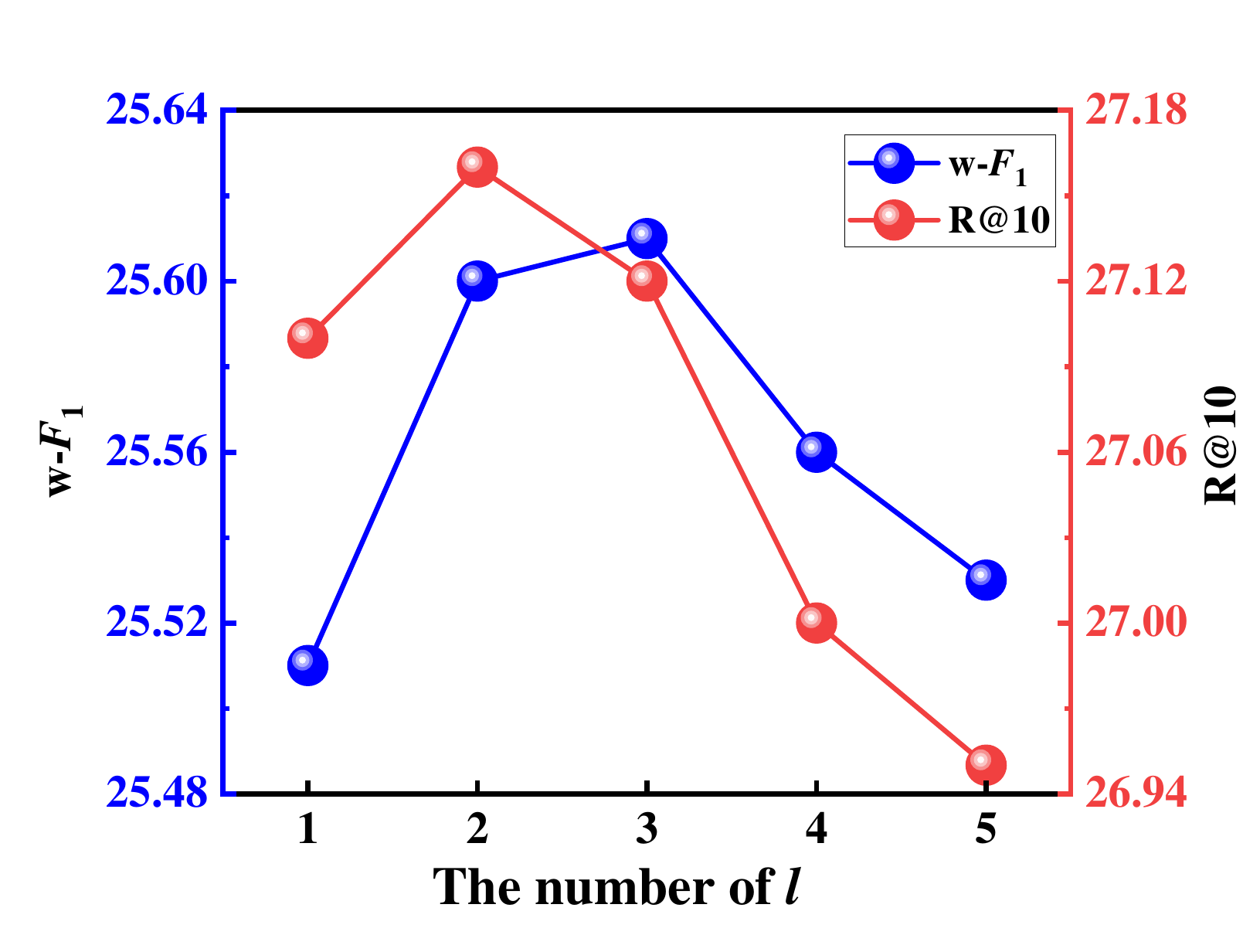}
		\caption{MIMIC-III dataset}
		\label{layer:mimic3}
	\end{subfigure}
    \hspace{-0.8em}
	\begin{subfigure}[b]{0.43\columnwidth}
		\centering
		\includegraphics[width=\linewidth]{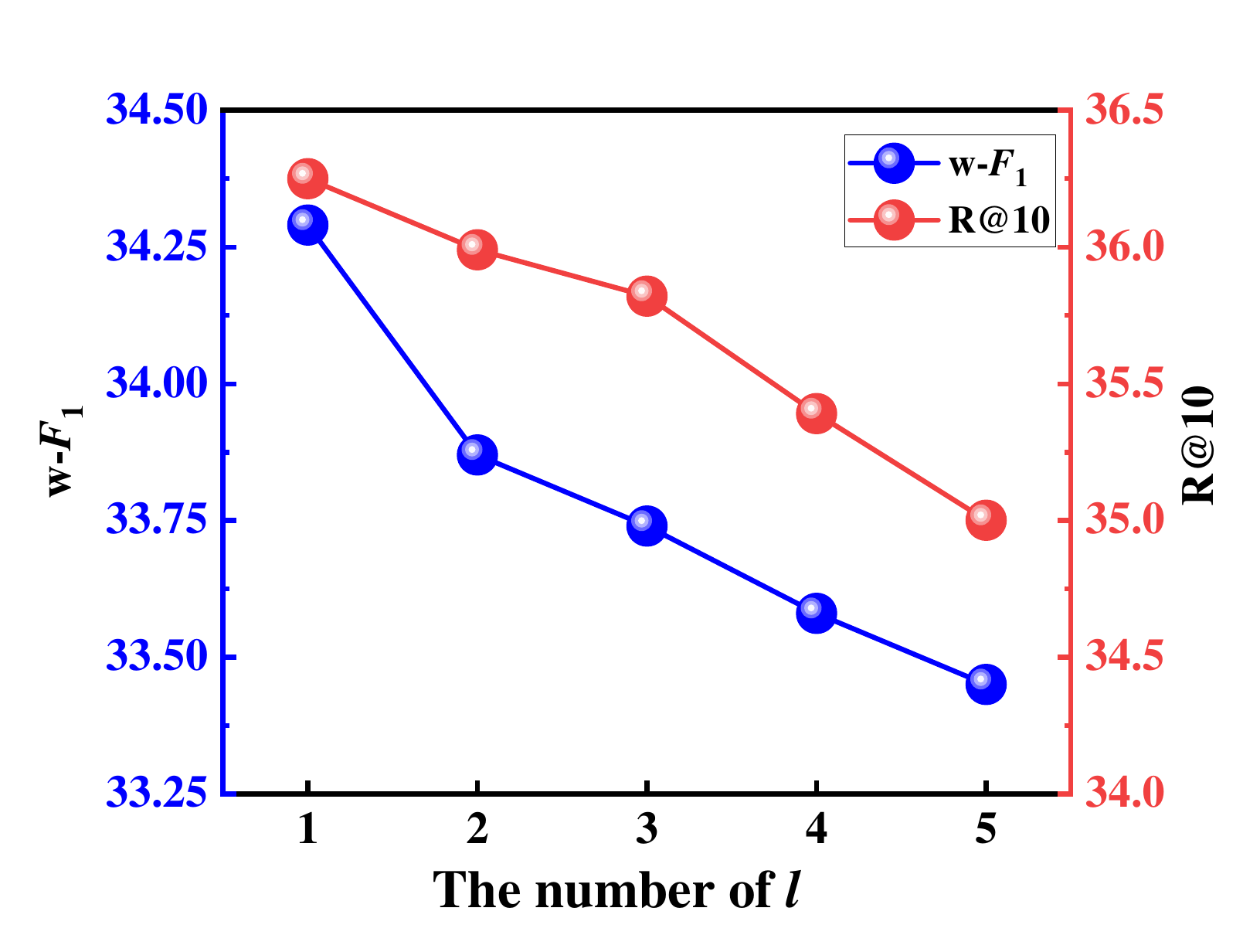}
		\caption{MIMIC-IV dataset}
		\label{layer:mimic4}
	\end{subfigure}
	\caption{Impact of different Transformer layers}
	\label{layer}
\end{figure}

\begin{figure}[!t]
	\centering
	\begin{subfigure}[b]{0.43\columnwidth}
		\centering
		\includegraphics[width=\linewidth]{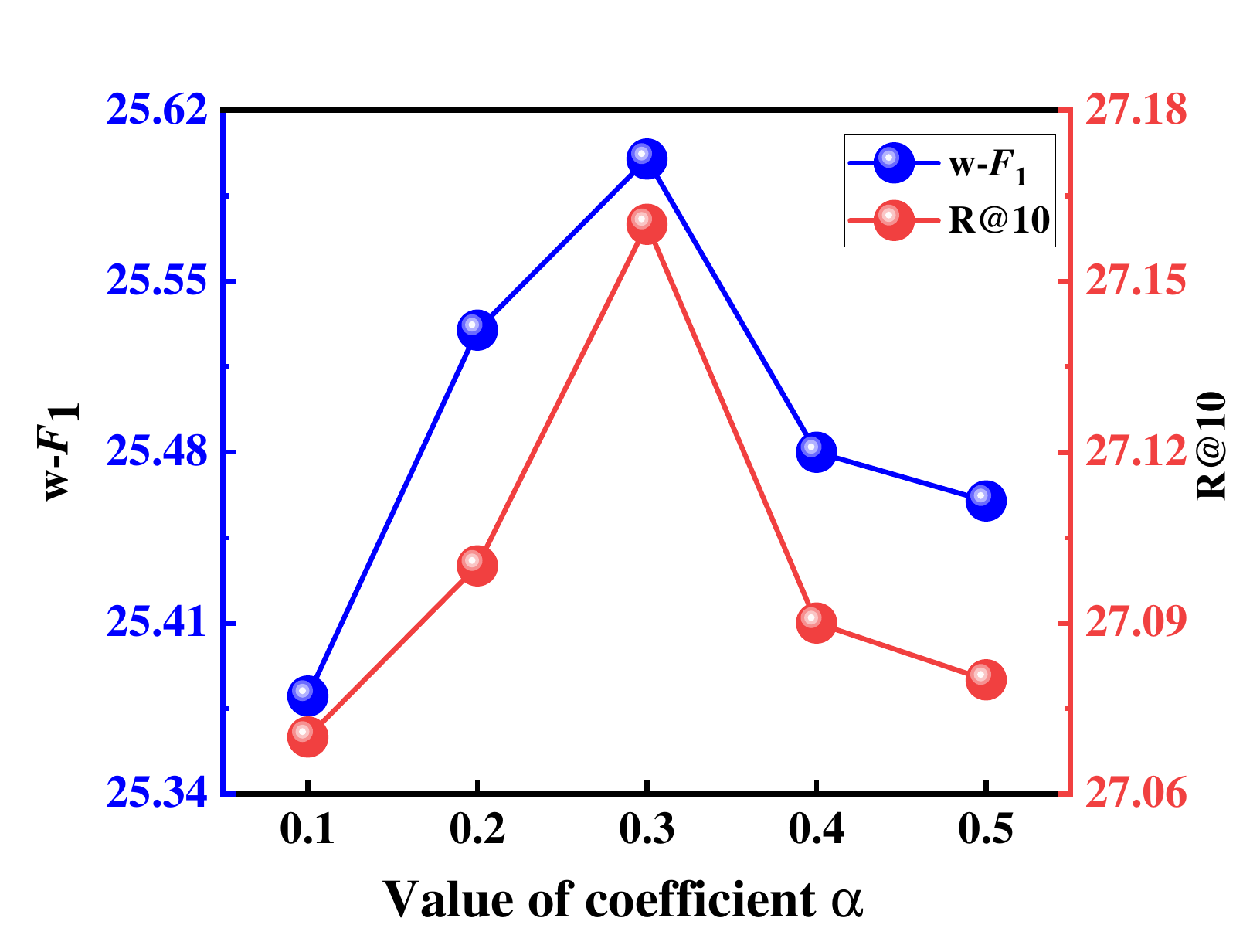}
		\caption{MIMIC-III dataset}
		\label{arpha:mimic3}
	\end{subfigure}
	\hspace{-0.8em}
	\begin{subfigure}[b]{0.43\columnwidth}
		\centering
		\includegraphics[width=\linewidth]{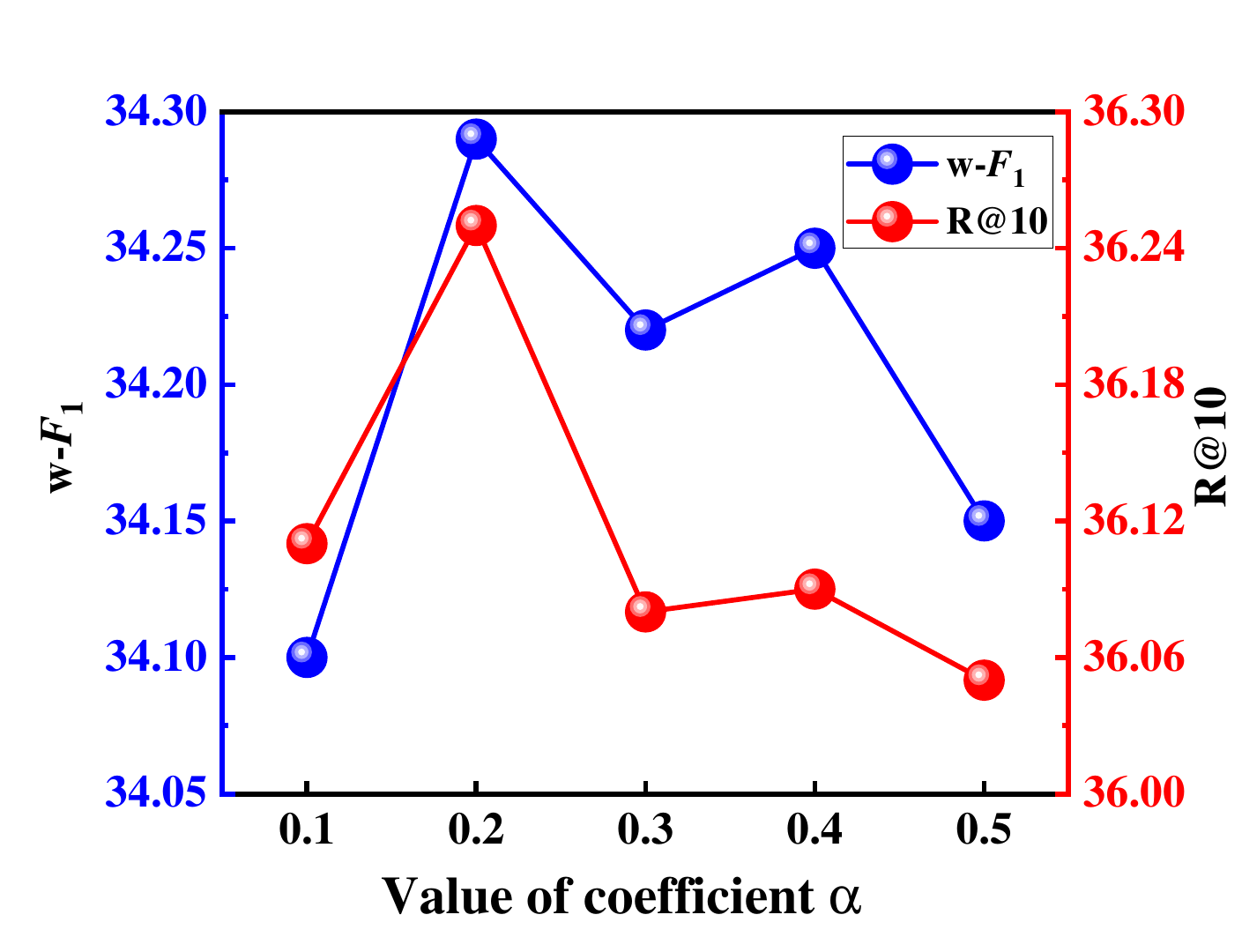}
		\caption{MIMIC-IV dataset}
		\label{arpha:mimic4}
	\end{subfigure}
	\caption{Impact of  hypergraph reconstruction loss coefficient}
	\label{loss}
\end{figure}

\subsection{Hyper-parameter Sensitivity Analysis}
To investigate the impact of key hyperparameters $\ell$ (number of Transformer layers) and $\alpha$ (hypergraph reconstruction loss coefficient) on the performance of SHGT, we conduct a series of controlled experiments. The results are reported using the w-$F_1$ and R@10 metrics,  with the corresponding plots presented in Fig.~\ref{layer} and Fig.~\ref{loss}.

Fig.~\ref{layer} presents the effect of varying the number of Transformer layers. On the MIMIC-III dataset, performance initially improves as $\ell$ increases,  peaking at $\ell = 2$, after which it begins to deteriorate as the number of layers further increases. In contrast, on the MIMIC-IV dataset, the model achieves the optimal result at $\ell = 1$, with further increases leading to performance degradation. These declines are likely attributable to overfitting caused by overly deep architectures, which may hinder the model's generalization ability.

Fig.~\ref{loss} illustrates the effect of the hypergraph reconstruction loss weight $\alpha$. Both datasets demonstrate a similar trend: performance improves with increasing $\alpha$, peaks at an optimal value, and then gradually declines. Specifically, the optimal values are $\alpha = 0.3$ for MIMIC-III and $\alpha = 0.2$ for MIMIC-IV. These findings suggest that carefully tuning $\alpha$ helps preserve the hypergraph's structure, enhancing predictive performance. However, an excessively large $\alpha$ may overemphasize structural constraints at the expense of semantic learning.

\section{CONCLUSION}
This work introduces SHGT, a novel structure-aware Hypergraph Transformer framework designed to advance diagnosis prediction accuracy. Unlike traditional graph-based approaches that capture only pairwise relationships, SHGT represents EHR data as a hypergraph, enabling effective modeling of higher-order interactions among medical codes. Subsequently, we employ a dedicated hypergraph structural encoder to extract rich local structural information. These structural features are then integrated into the Transformer, facilitating comprehensive modeling of global code dependencies across the entire hypergraph. To further preserve structural semantics and enhance embedding quality, we incorporate a hypergraph reconstruction loss that ensures alignment between code and visit embeddings. Extensive experiments conducted on two real-world EHR datasets demonstrate the superior performance of the proposed framework. In future work, we plan to integrate hypergraph neural networks with large language models to generate informative code embeddings and further improve diagnostic accuracy.

\bibliographystyle{IEEEtran} 
\bibliography{references}    

\end{document}